\documentclass[runningheads]{llncs}
\usepackage{eccv}

\usepackage{eccvabbrv}

\usepackage{graphicx}
\usepackage{booktabs}
\usepackage{multirow}
\usepackage{tabularx}

\definecolor{GBColor}{rgb}{0.0,0.0,1.0} 

\definecolor{GBRColor}{rgb}{1.0,0.0,1.0} 

\usepackage[accsupp]{axessibility}  

\usepackage{hyperref}

\usepackage{orcidlink}

\begin{document}

\title{StyleTokenizer: Defining Image Style by a Single Instance for Controlling Diffusion Models} 

\titlerunning{StyleTokenizer}



\newcommand{\equalcontrib}{\textsuperscript{\dag}}
\footnotetext[1]{\equalcontrib Equal contribution}

\author{
Wen Li\inst{}\equalcontrib \and
Muyuan Fang\inst{}\equalcontrib \and
Cheng Zou\inst{} \and
Biao Gong\inst{} \and
Ruobing Zheng\inst{} \and \\
Meng Wang\inst{} \and
Jingdong Chen\inst{} \and
Ming Yang\inst{}
}

\authorrunning{W.~Author et al.}

\institute{Ant Group, Hangzhou, China \\
\email{\{liwen8459,fangmuyuan\}@gmail.com \{wuyou.zc,gongbiao.gb,\\zhengruobing.zrb,darren.wm,jingdongchen.cjd,m.yang\}@antgroup.com}}

\maketitle
\begin{abstract}

Despite the burst of innovative methods for controlling the diffusion process, effectively controlling image styles in text-to-image generation remains a challenging task. Many adapter-based methods impose image representation conditions on the denoising process to accomplish image control. However these conditions are not aligned with the word embedding space, leading to interference between image and text control conditions and the potential loss of semantic information from the text prompt. Addressing this issue involves two key challenges. Firstly, how to inject the style representation without compromising the effectiveness of text representation in control. Secondly, how to obtain the accurate style representation from a \emph{single} reference image. To tackle these challenges, we introduce StyleTokenizer, a zero-shot style control image generation method that aligns style representation with text representation using a style tokenizer. This alignment effectively minimizes the impact on the effectiveness of text prompts. Furthermore, we collect a well-labeled style dataset named Style30k to train a style feature extractor capable of accurately representing style while excluding other content information. Experimental results demonstrate that our method fully grasps the style characteristics of the reference image, generating appealing images that are consistent with both the target image style and text prompt. The code and dataset are available at \href{https://github.com/alipay/style-tokenizer}{https://github.com/alipay/style-tokenizer}.

\end{abstract}   
\begin{figure}[!h]
	\centering
	\includegraphics[width=1\textwidth]{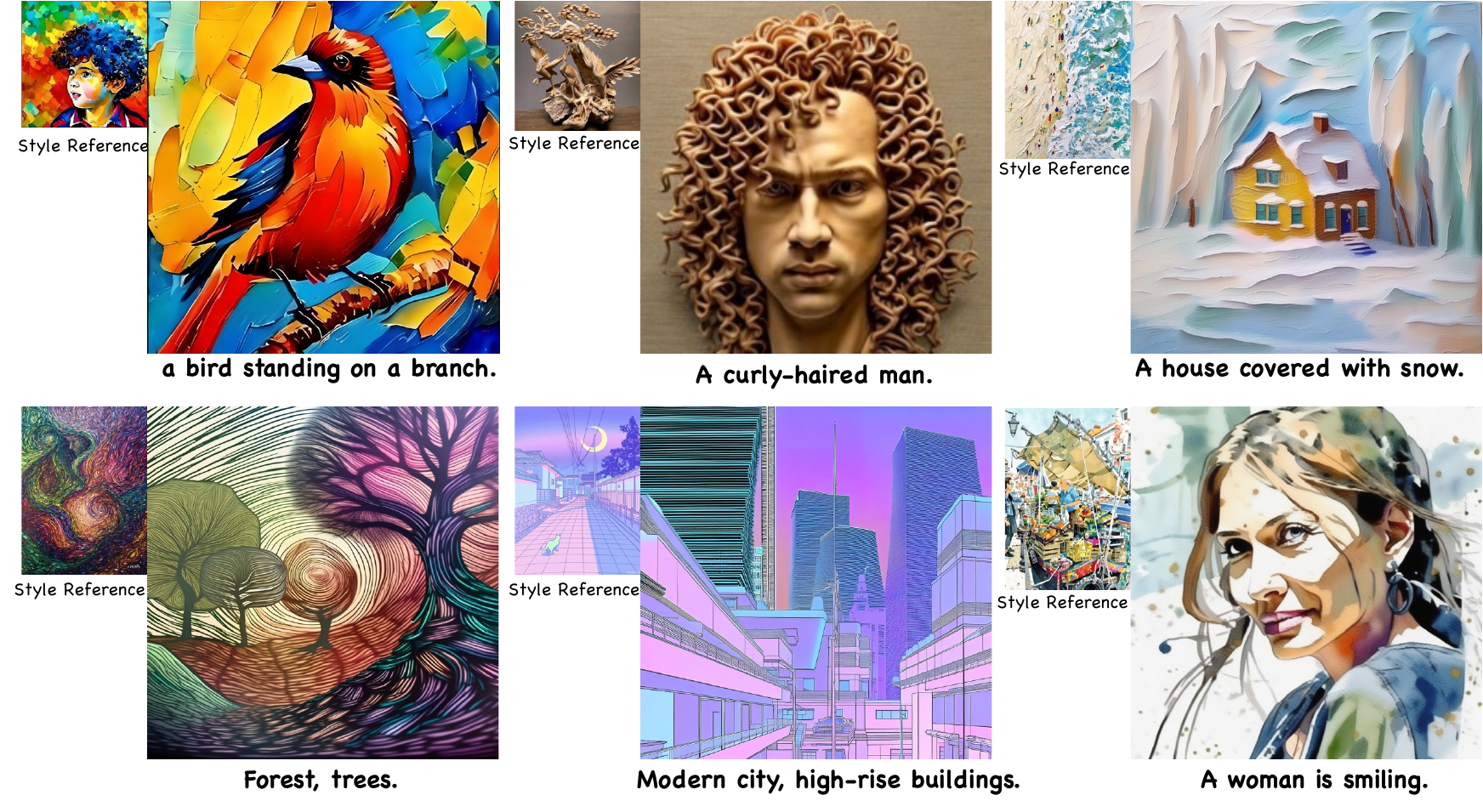}
	\caption{\textbf{Some showcases of StyleTokenizer.} It is capable of generating images with corresponding styles using a single style image reference. For each image pair, the smaller one is a style reference, and the larger one is a generated image conditioned by the corresponding style reference and text prompt on the bottom.}
	\label{fig:impressive_case}
\end{figure}

\section{Introduction}
\label{sec:intro}

\begin{figure}[tb]
	\centering
	\includegraphics[width=1\textwidth]{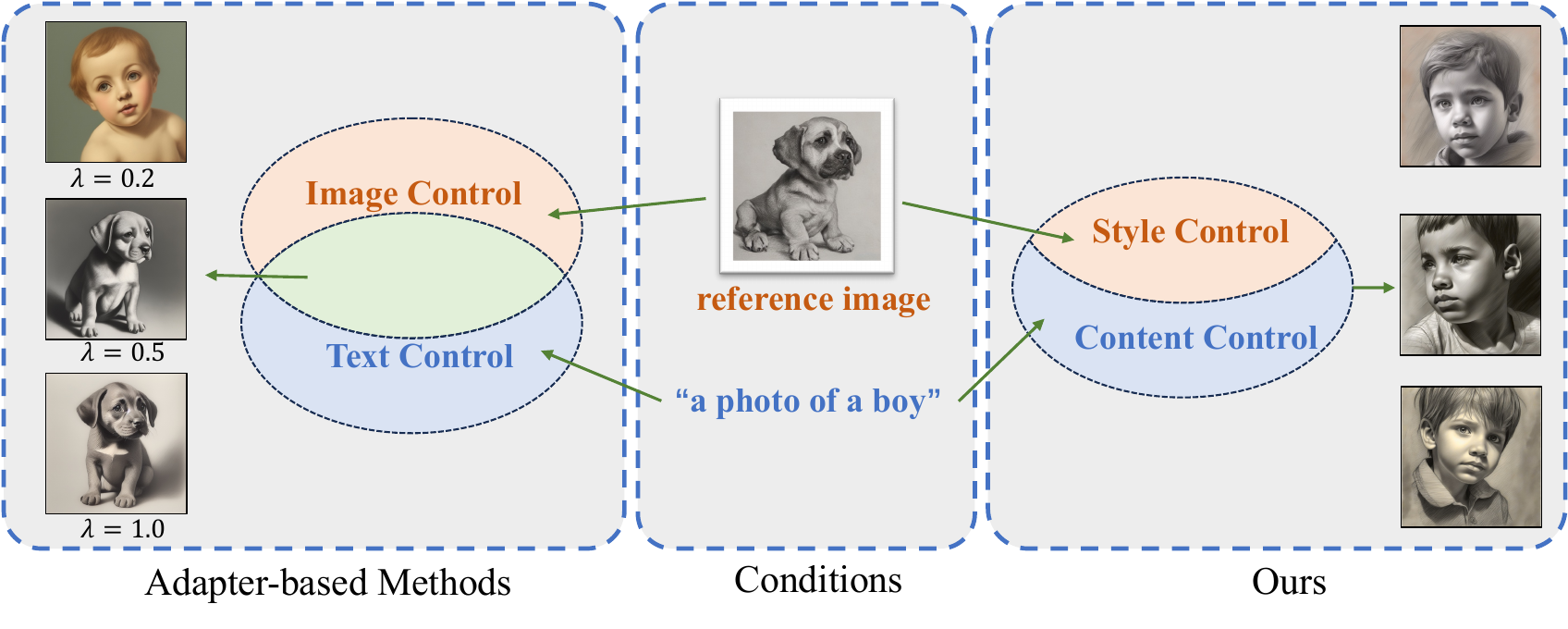}
	\caption{\textbf{The difference between with adapter-based methods.} \emph{Left:} Adapter-based methods directly inject image representation in a similar manner with text representation, leading to interference between the two control conditions and loss of semantic information from the text prompt. \emph{Right:} StyleTokenizer aligns style representation with text representation into a common semantic space, which minimizes the impact on the effectiveness of text prompts. }
	\label{fig:method_diff}
\end{figure}

The field of image generation has experienced remarkable growth since the advent of diffusion-based methods, including notable examples such as DALLE-1/2/3~\cite{ramesh2021zero, ramesh2022hierarchical, dalle3}, Stable Diffusion~\cite{rombach2022high}, and Midjourney~\cite{midjourney}. These advancements have paved the way for a diverse range of content control techniques~\cite{zhang2023adding,mou2023t2i}, enabling precise manipulation of layout, lines, depth, and other conditions. This not only enhances the stability of diffusion models but also broadens their applicability.
Despite such progress, achieving effortless and effective control over the fine-grained styles of synthesized images remains a formidable challenge. This limitation restricts the practical applicability and convenience of diffusion methods in various applications.

Previous GAN-based methods~\cite{zhu2017unpaired,choi2020stargan,huang2017arbitrary,deng2022stytr2,liu2021blendgan} have achieved some level of style control, but the generated effects are difficult to compare with those of diffusion models.
Diffusion-based methods such as Textual Inversion~\cite{gal2022image}, LoRA~\cite{hu2021lora}, and Dreambooth~\cite{ruiz2023dreambooth} utilize a small amount of data of the same type to fine-tune pre-trained
text-to-image models to better reflect new aspects in training images. These methods can generate images with similar styles as the training images. However, they are also prone to overfitting with the specific content (e.g., a particular person or object) present in the training images. This makes it challenging to decouple style and content, resulting in difficulties in achieving precise style control.

For precise style control, one intuitive approach is to employ adapter-based techniques like IP-adapter~\cite{ye2023ip-adapter}, 
 StyleAdapter~\cite{wang2023styleadapter}, InstantStyle~\cite{wang2024instantstyle}, etc. These strategies embed style representation within the UNet architecture by introducing an extra cross-attention layer. Yet, as text and style representations span in distinct spaces, managing them via individual cross-attention processes may lead to discrepancies in the control signals.
As illustrated in Fig.\ref{fig:method_diff}, the adapter-based approaches apply text and style conditions simultaneously during the denoising process, which may cause interference between the controls and losing semantics.
Thus, the crucial challenge lies in introducing style representation while preserving the integrity of text representation for control purposes.

Leveraging the approach of tokenizing visual features for alignment with linguistic space, such as LLaVA~\cite{liu2023llava}, we can improve the handling of intricate details in both images and text. Specifically, tokenizing style elements like text prompts significantly enhances the coordination and control within text prompts. This approach also effectively decouples style and content representations: after tokenization, the style from reference images and content extracted from text prompts remain in their distinct semantic spaces without overlap. Consequently, in this way, we can simultaneously achieve style and content control during generation without any interference.

Another challenging aspect is how to obtain accurate style representation from a single reference image, as most existing methods~\cite{2023i2vgenxl, ye2023ip-adapter} simply using the coarse-grained supervision trained CLIP encoder struggle with independent style control.
To address this, we have developed a style-focused dataset with over 300 style categories and 30,000 images, all professionally annotated. 
In addition, based on the tokenization method we analyzed in the previous paragraph for precise control and decoupling, we have trained a style-specific embedding extractor, enhanced by contrastive learning, to distinguish and represent style nuances. This refinement boosts the encoder's adaptability to new styles and overall robustness.

Our contributions can be summarized as follows: 
\begin{itemize}
\item[$\bullet$] We introduce StyleTokenizer, a novel method for style control in diffusion models. This approach allows for accurate style control of generated images using a single arbitrary reference image in a training-free manner, while minimizing the impact on the control effectiveness of text prompts. Experimental results demonstrate the outstanding performance of our proposed method compared to other state-of-the-art approaches in this field.
\item[$\bullet$] We curate a Style30k dataset comprising over 300 widely distributed style categories, manually collected by professional designers. This dataset includes a total of 30,000 images and, to the best of our knowledge, is currently the largest and most diverse open-source style dataset available. Using this dataset, we train a robust style encoder capable of effectively representing style information based on a single reference image.
\end{itemize}

\section{Related Works}
\label{sec:formatting}

\subsection{Text-to-Image synthesis}

Text-to-image synthesis has experienced a phenomenal technological breakthrough in recent years.
DALL-E 1, 2, 3~\cite{ramesh2021zero,ramesh2022hierarchical,dalle3} has demonstrated impressive results in text-to-image synthesis by utilizing a text encoder to control auto-regressive transformers and diffusion models. This led to substantially refined and high-visual-quality synthetic images. Stable Diffusion~\cite{rombach2022high} and Imagen~\cite{saharia2022photorealistic} have also shown promising results in text-to-image synthesis by leveraging diffusion models.
Furthermore, StyleGAN-T~\cite{sauer2023stylegan} has explored the potential of GANs in text-to-image synthesis and demonstrated remarkable results. These approaches typically involve using text encoders like CLIP~\cite{radford2021learning} and GPT~\cite{brown2020language} and subsequently controlling the generators.
There are many attribute control methods derived from text-to-image generation models. ControlNet~\cite{zhang2023adding} incorporated additional control features into Unet's feature space. Composer~\cite{huang2023composer} added extra feature inputs to cross attention and time embedding. Text inversion~\cite{zhang2023inversion} and blip-diffusion~\cite{li2023blip}, on the other hand, controlled attributes by introducing novel text embeddings. These methods have greatly advanced the field of text-to-image synthesis.

\subsection{Style Control in Image Generation}

Style is a quite subtle attribute of images that is hard to define or even describe by people. 
Even though style control has been widely adopted in diverse applications, such as design and short video entertainment, it involves considerable efforts including scaffolding tuning and try-and-error from users. Gatys~\cite{gatys2015neural} proposed the use of VGG and Gram matrix to extract style features for loss supervision. CycleGAN~\cite{zhu2017unpaired} and StarGAN~\cite{choi2018stargan,choi2020stargan} established cycle mechanisms to achieve style transfer with a limited set of styles while keeping the content intact. BlendGAN~\cite{liu2021blendgan} employed an additional style encoder to control StyleGAN~\cite{karras2020analyzing} for zero-shot style transfer on face datasets. CAST~\cite{zhang2022domain} utilized a contrastive loss to extract image styles. ALADIN~\cite{ruta2021aladin} used the AdaIN~\cite{huang2017arbitrary} module to extract styles and control the decoder for style image generation. Wu~\cite{wu2023not} attempted to extract style information from visual features by aligning the generated images from diffusion with the style and content components of the prompt. 
Some methods leverage prior knowledge by assuming that styles mainly reside in certain feature categories to achieve style control. DiffStyle~\cite{huang2022diffstyler} combined diffusion models of both style and content during denoising. ProsPect~\cite{zhang2023prospect} incorporated content and style reference images at different denoising stages in the diffusion process. P+~\cite{voynov2023p+} controlled the features of Unet at different resolutions separately for content and style. SDXL~\cite{podell2023sdxl} incorporated an additional prompt module that enables users to specify limited styles. Nonetheless, these prior arts still require some costly efforts of data collection or finetuning from the end user. This limitation has been resolved by our method, which involves extracting style and applying control from a single image.

\section{Methodology}

\begin{figure*}[tb]
	\centering
	\includegraphics[width=1\textwidth]{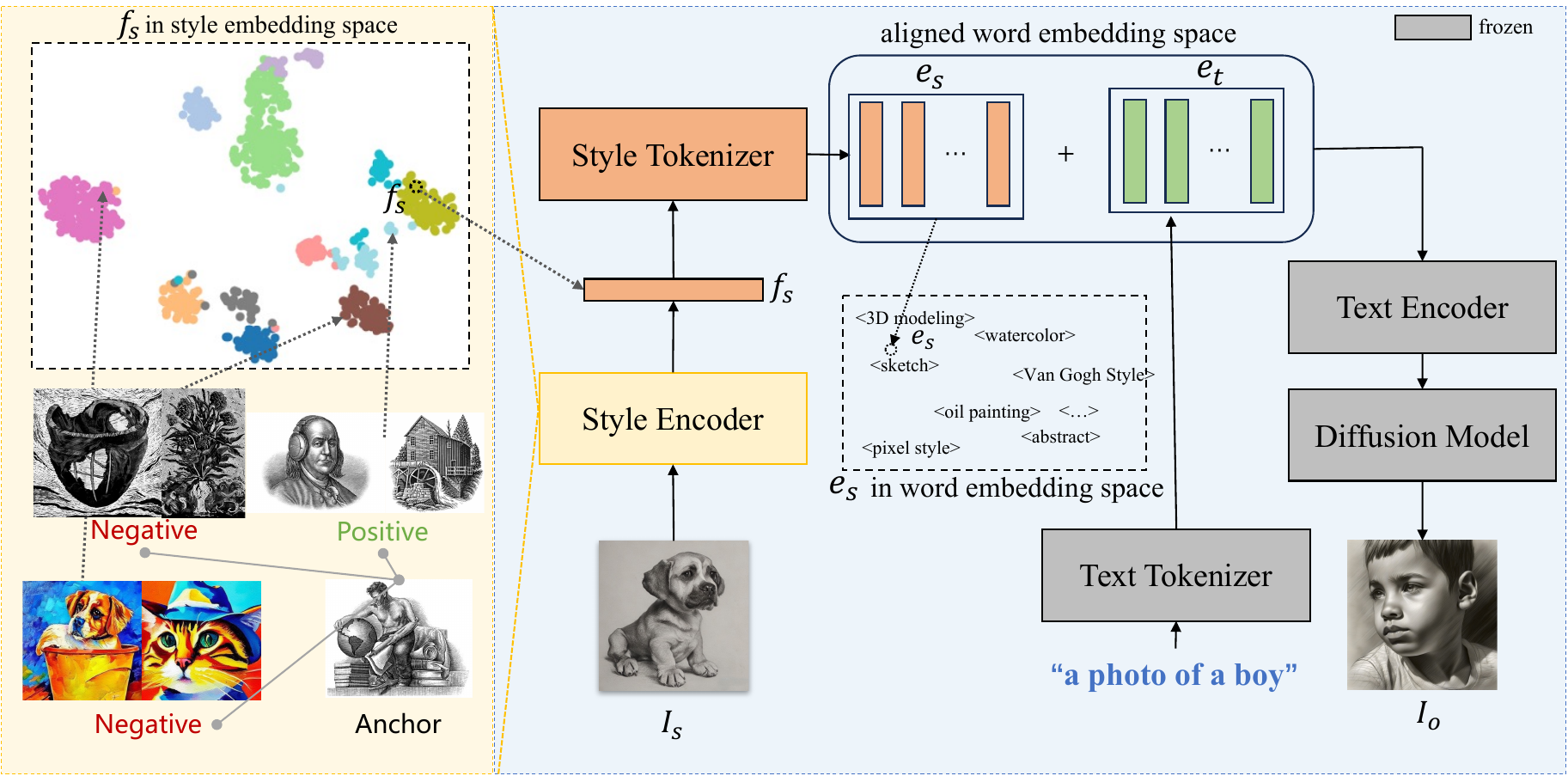}
	\caption{\textbf{Overview of StyleTokenizer.} Our method consists of two stages. In the first stage, a Style Encoder is trained on a style dataset to acquire style representation capabilities. We employ contrastive learning to enforce it to focus on the distance differences between diverse styles for better style representation. In the second stage, style embedding is extracted from a single image by a Style Encoder, and then a Style Tokenizer converts it into style tokens, which are aligned with text tokens in the word embedding space. Finally, these tokens are input to the SD pipeline as a condition to generate the image.}
 
	\label{fig:pipeline}
\end{figure*}

\subsection{Overview}

Our method is derived within the Stable Diffusion framework, which decouples content and style conditions in the image generation process, resulting in visually appealing and coherent outputs. In this section, we introduce the overall pipeline of our method. 

Compared with the traditional Stable Diffusion framework, we introduce two novel modules, as illustrated in Fig.~\ref{fig:pipeline}, a Style Encoder for style representation and a Style Tokenizer for style control. These two modules are trained in two stages. In the first stage, the Style Encoder is trained on a style dataset named Style30K to acquire style representation capability. In the second stage, style representations are extracted from reference images by Style Encoder, and Style Tokenizer learns to convert them into style tokens, which are aligned with text tokens in the word embedding space. Finally, both text tokens and style tokens are concatenated and input to the SD pipeline as a condition to generate the image.

\subsection{Style30K Dataset}
\label{sec:style30k}
\begin{figure}[tb]
	\centering
	\includegraphics[width=1\textwidth]{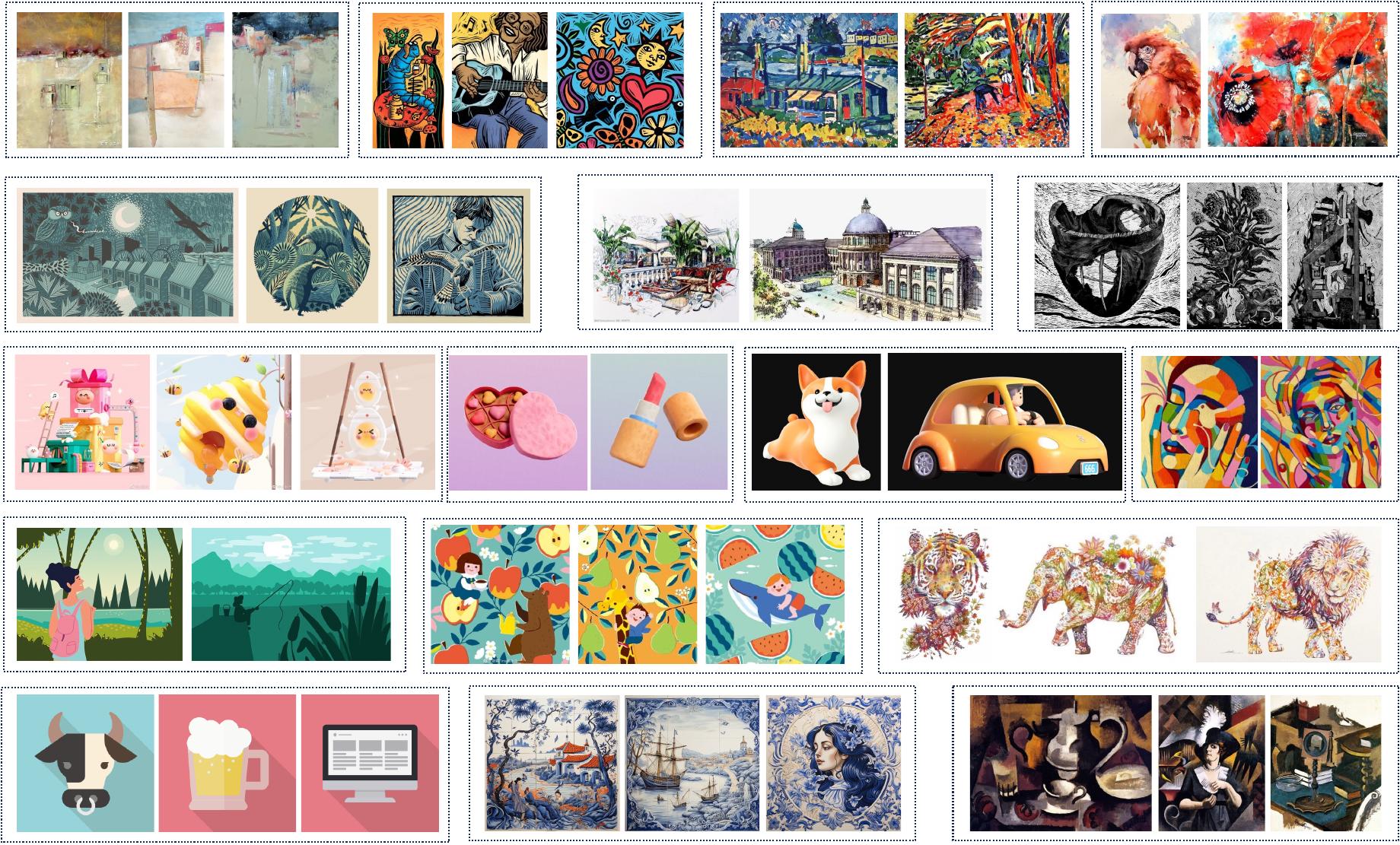}
	\caption{\textbf{Partial style images in Style30k dataset.} Each dotted box represents a style category. The Style30K is a style-focused dataset with over 300 style categories and 30,000 images, all professionally annotated. The number of images in each category ranges from 30 to 200. These categories cover a diverse range of fields, including art styles, commercial design styles, 3D modeling, \emph{etc.}}
	\label{fig:dataset}
\end{figure}

Describing the style of an image verbally is even challenging for artists. The meaning of image style is very rich and subtle, encompassing various perspectives, such as color distribution, lighting, line styles, artistic styles, brushwork, emotions, \emph{etc.} These characteristics are visually perceptible but difficult to precisely and comprehensively describe in language. Therefore, extracting features directly from images is a more reasonable choice, rather than relying on text. Existing feature extraction methods often rely on generic semantics or classification features and are not specifically trained for style-related tasks. Hence, we construct a style dataset named Style30K to train a style encoder that focuses on capturing style-oriented features, shown in Fig.~\ref{fig:dataset}.

While it is challenging to describe image styles in language, people can intuitively judge whether two images share the same style or not. Therefore, we adopt a semi-manual approach to construct the dataset. The collection process of Style30K consists of three stages. In the first stage, we gather images with various styles, where each style is represented by three sample images as queries. Subsequently, we use different embedding extractors to extract the embedding of these queries and perform retrieval within a large dataset. In the second stage, we manually filter and collect images from the retrieval results that share the same style as the three queries for each style category. Each collected image requires a consensus among three annotators for it to be included. In the third stage, we annotate each collected image with a content prompt using CogVLM~\cite{wang2023cogvlm}. In detail, the image and the instruction of "Describe the image, only the content of the image, do not include the style of it." are input into the caption model, which yields captions that are solely related to the content of the image. The reason for this way is to ensure that style and content control signals are independent to each other. During the model's training, the prompt provides information related only to content, while the style is provided by the Style Tokenizer.

\subsection{Style Encoder}
In this section, we describe the process of training the Style Encoder $\mathcal{E}_s$, which extracts the style cue from images $I_s$ and encodes it into style embedding $f_s$. This embedding is then used to guide the generation process. 

Obtaining accurate style representation from a single reference image is a challenging task. Previous methods extract image representation from the CLIP image encoder to enable content and style control. This approach has shown promising results in providing effective control over various visual aspects, including color, subject, layout, and more. However, it lacks the ability to independently control these aspects, particularly style control, as CLIP is trained using coarse-grained semantic information as supervision. 

To address this limitation, we use the well-labeled style dataset Style30K to train a style representation extractor capable of accurately representing style while excluding other content information. Given that the style data includes an accurate category label, the style encoder is trained using supervised representation learning. It enforces the model to only focus on category-related information (style) and ignore category-irrelevant information (content). To further enhance the generalization of the style encoder, we employ contrastive loss for supervision. As shown in the left of Fig.\ref{fig:pipeline}, it allows the model to focus on the distance differences between diverse styles. Different images of the same style are clustered together in the embedding space, images of similar styles are closer, while images of different styles are scattered. This approach enhances the robustness of the style encoder in processing new styles and improving its ability to handle novel style variations.

\subsection{Style Control}

Previous adapter-based methods have the capability of image prompting in the diffusion model. It significantly enhances the ability to generate content that cannot be described in a prompt. These methods incorporate style representation into the Unet module using an additional cross-attention layer. However, they apply text and style conditions simultaneously during the denoising process, which may cause interference between the controls and loss of semantics.

Representations from the word embedding space of SD have rich style control capabilities. On the one hand, Dreambooth~\cite{ruiz2023dreambooth} and Textual Inversion~\cite{zhang2023inversion} have demonstrated that new word embedding outside the existing dictionary can express diverse contents. Yet, these methods require some reference images for tuning and are prone to overfitting with the specific content. On the other hand, carefully crafted descriptions by prompt experts can also yield desired image styles. However, directly using textual descriptions to control the style is still challenging. The text descriptions used during the training of SD lack a detailed description of the style for each image. Besides, image styles encompass a wide range of aspects that are difficult to fully express in natural language. 

Therefore, we aim to find a comprehensive and accurate style description that can be applied to each image and is acceptable by diffusion pipelines. Considering that our Style Encoder is already capable of extracting a unique style embedding for any given image, a reasonable approach is to map these styles to representations in the space of word embedding. We utilize a 2-layer MLP named Style Tokenizer $T_s$ to implement this mapping. The Style Tokenizer $T_s$ takes the embedding $f_s$ extracted by the Style Encoder $\mathcal{E}_s$ and maps them into style embedding tokens $e_s$. In the training process, the parameters of the original SD model are frozen, and only the parameters of Style Tokenizer are updated, enabling the mapped embedding $e_s$ to provide a comprehensive and precise representation of image styles. The style embedding $e_s$ is concatenated with word embedding tokens $e_t$ following Eq.~\ref{eq:cat_token} and then fed into SD's text encoder. By doing so, style images can be used as a style prompt when generating images to better describe the style. Besides, the style from reference images and content extracted from text prompts remain in their distinct semantic spaces without overlap. 

\begin{equation}
    e_{ts} = [e_{start},e_s,e_t,e_{end}],
    \label{eq:cat_token}
\end{equation}
where $e_s = T_s(f_s)$.

During the inference process, we employ the classifier-free guidance~\cite{ho2022classifier}. In order to independently control the weight of both text and style, we adopt a similar approach as InstructPix2Pix~\cite{brooks2023instructpix2pix}, as described in Eq.~\ref{eq:cfg}. 

\begin{equation}
    \begin{aligned}
        \tilde{e}_\theta(z_t, c_t, c_s) &= e_\theta(z_t, \emptyset, \emptyset) \\
        & + s_t \cdot (e_\theta(z_t, c_t, \emptyset) - e_\theta(z_t, \emptyset, \emptyset)) \\
        & + s_s \cdot (e_\theta(z_t, c_t, c_s) - e_\theta(z_t, c_t, \emptyset)),
        \label{eq:cfg}
    \end{aligned}
\end{equation}
where $c_t$ and $c_s$ represent the text and style condition. The strength of $s_t$ and $s_s$ represents the intensity of the text and style condition, respectively.

\section{Experiments}

\subsection{Experimental Details}
\label{sec:exp}

We collect 10 million high-quality images on the Internet for model training. The Style Encoder adopts the visual encoder of CLIP ViT-B/32 as the pre-trained backbone. We first use these 10 million images for pre-training, and then use Style30K for supervised training. In terms of style control, we adopt SD v1.5 as the generation model. The style embedding is tokenized by the Style Tokenizer with a shape of $8\times768$. Then, these tokens are concatenated with text embedding tokens and fed into the text encoder of the SD pipeline.
During this stage, all the images are used for training the Style Tokenizer. When the denoising target image $I$ is from the Style30K dataset, the style reference image $I_s$ is randomly selected from the images within the same style category as $I$. Otherwise, when $I$ does not have style annotation, the style reference is $I$ itself.

\begin{figure*}[!t]
	\centering
	\includegraphics[width=1\textwidth]{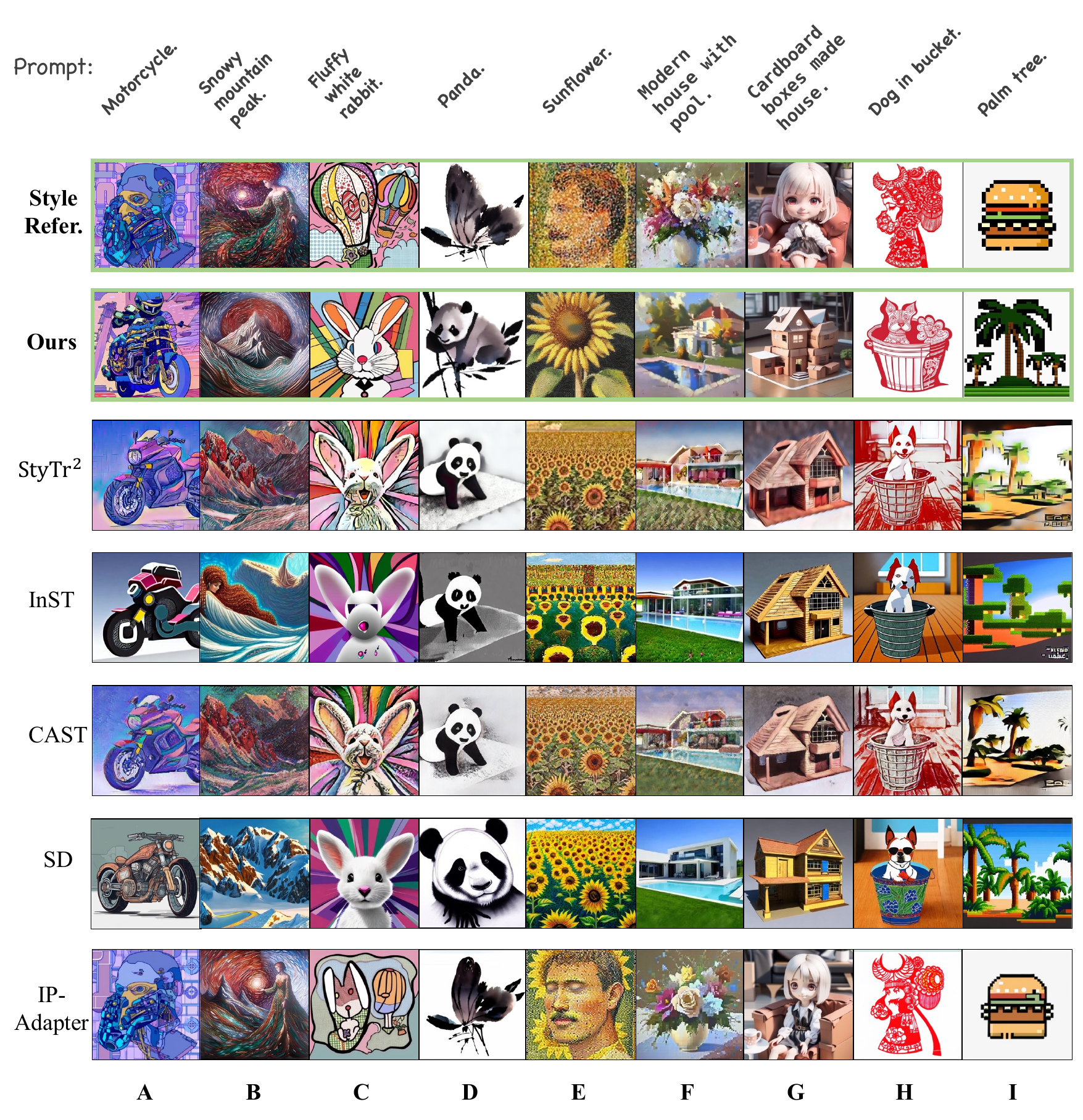}
	\caption{\textbf{Visual comparison with the competing methods.} Each column represents the results generated by different methods using the same prompt and reference image. Each row represents the result generated by a different method. Images in the first row are style references used to control style.}
	\label{fig:sota}
\end{figure*}

\subsection{Qualitative Evaluation}
\label{sec:qualitative_evaluation}
To facilitate a comparison with other methods, we conducted a test on both our method and the previous approaches, including $StyTr^2$\cite{deng2022stytr2}, InST\cite{zhang2023inversion}, CAST\cite{huang2017arbitrary}, SD\cite{rombach2022high} controlled by style prompt, and IP-Adapter\cite{ye2023ip-adapter}. To evaluate the performance of these methods in both style control and prompt following, we prepare a benchmark consisting of 52 prompts and 28 style reference images, these prompts are from the settings used in the StyleAdapter~\cite{wang2023styleadapter}. The prompts encompass a rich variety of content, including human, animal, object, and scene. The reference images cover some common styles as well as some that are difficult to describe in words. Note that both of them are excluded from the training process. Our aim with the aforementioned setup is to comprehensively evaluate the strengths and weaknesses of the different methods. Some images generated by these methods can be viewed in Fig.\ref{fig:sota}, where each column represents the results produced by different methods using the same prompt and reference image. Below, we provide a detailed analysis of the experimental results.

As shown in Fig.\ref{fig:sota}, $StyTr^2$ and InST achieve relatively similar performance, successfully capturing the dominant color palette of the reference images. However, their grasp of the overall style such as texture is not very well. As seen in column $H$, they capture the red color information from the reference image but fail to comprehend the cut-paper style. Furthermore, their image quality is generally inferior to that of other methods. Utilizing style prompts for control facilitates a certain level of style control in simpler style categories, like oil and ink wash paintings, but the absence of a reference image as input led to significant discrepancies in the finer details. For more complex styles that are difficult to articulate, their ability to control style is lost. IP-Adapter produces images with a style very close to the original, but in most cases, it struggles to decouple the content from the reference image, leading to poor prompt-following ability. For instance, in columns $B$ and $E$, although mountains and sunflowers are generated, the human from the reference images also appear in the output. IP-Adapter's strengths are mainly in image variation and image editing. In contrast, our method demonstrates a high degree of consistency with the reference image in terms of style, including line, texture, color, emotion, and more. It also shows a strong advantage in following text prompts. Moreover, the overall aesthetic quality of the images is superior to that of the previous methods.

\subsection{Quantitative Evaluation}
\label{sec:sota}


\begin{table}[tb]
    \centering
    \caption{\textbf{Comparison of different methods using quantitative evaluation.} }
    \label{tab:metrics}
    \setlength{\tabcolsep}{4mm}{
    \begin{tabularx}{\linewidth}{c|ccccc}
    \hline
    \multirow{2}{*}{Methods} & \multicolumn{3}{c}{Model-based Metric}          & \multicolumn{2}{c}{User Study ($\%$)} \\
                             & \textbf{Style-Sim}      & Aesthetic     & Text-Sim       & Pro.  & Ord.  \\
    \hline
    Cast~\cite{huang2017arbitrary}                     & 0.364          & 6.05          & 0.272          & 19.5          & 14.6            \\
    $StyTr^2$~\cite{deng2022stytr2}                    & 0.443          & 5.92          & 0.262          & 6.2           & 6.4            \\
    InST~\cite{zhang2023inversion}                     & 0.190          & 5.84          & 0.242          & 12.5           & 10.7            \\
    SD~\cite{rombach2022high}                       & 0.178          & 6.55          & \textbf{0.279} & /             & /              \\
    IP-Adapter~\cite{ye2023ip-adapter}                       & 0.480          & 6.59          & 0.116 & /             & /              \\
    \textbf{Ours}                     & \textbf{0.482} & \textbf{6.68} & 0.277          & \textbf{61.8} & \textbf{68.3}            \\
    \hline
    \end{tabularx}}
\end{table}

In this section, our method is compared with the state-of-the-art approaches using some quantitative metrics for a fair evaluation. We employ the following metrics on the images generated in the Sec.\ref{sec:qualitative_evaluation} to evaluate the quality and effectiveness of the generated images:

\noindent\textbf{Text-Image Similarity.} We use the CLIP model to extract embedding from generated images and their corresponding text prompts. Then cosine similarity between embedding from the prompt and the generated image is calculated. Higher cosine similarity indicates better capability of the instruction following.

\noindent\textbf{Aesthetic Score.}
To assess the aesthetic quality of the generated images, we predict the aesthetic score for each generated image with LAION-Aesthetics Predictor~\cite{laionaesthetics}. This metric measures the visual appeal and artistic quality of the image. A higher aesthetic score indicates a visually pleasing image.

\noindent\textbf{Style Similarity.}
Since there is no generally accepted method for assessing style similarity, we imitate the text-image similarity metric calculated by CLIP. Style embedding of both the style reference image and the generated image are exacted by Style Encoder. After that, we compute their cosine similarity. A higher cosine similarity indicates better control of the desired style in the generated images.

\noindent\textbf{User Study.}
To assess the style similarity more comprehensively, we conduct user studies. For the generated images produced by each method, we have 20 users (10 professional designers and 10 users) vote anonymously for the image that they believe has the most similar style to the reference image. 
The normalized votes (vote rate), serve as the style similarity score.

For each of these metrics, we calculate the average across all generated results to provide an overall evaluation of the performance of the existing style control models. Experimental results are summarized in Tab.~\ref{tab:metrics}, where we compare our method with the SOTA approaches using the aforementioned evaluation metrics. 
Our method significantly outperforms other SOTA approaches in terms of style similarity. In the user study, our method also receives more votes than other methods. These results highlight the effectiveness of our approach in preserving the desired style in the generated images. Furthermore, our method is trained on large-scale high-aesthetic data and thus achieves a higher aesthetic score than the base SD model. As shown in Fig.\ref{fig:sota}, it brings better results in terms of aesthetics than other methods. As for the instruction following, the text-image similarity of our method has comparable performance with the base SD model. It shows that our method does not lead to a decrease in the ability to follow instructions during style control. In summary, the experimental results demonstrate that our method can achieve better style control capability and generate visually appealing images, while the instruction following is not affected by this.

\subsection{Evaluation of Style Encoder}

We conduct an evaluation of our Style Encoder and compare it with several publicly available feature encoders, namely CLIP~\cite{radford2021learning}, VGG~\cite{simonyan2014very} and BlendGAN~\cite{liu2021blendgan}. The evaluation is performed on a validation set of Style30K, consisting of 12 different style categories with a total of 900 images that are non-repetitive with the training style categories.

\begin{figure}[tb]
	\centering
	\includegraphics[width=0.7\textwidth]{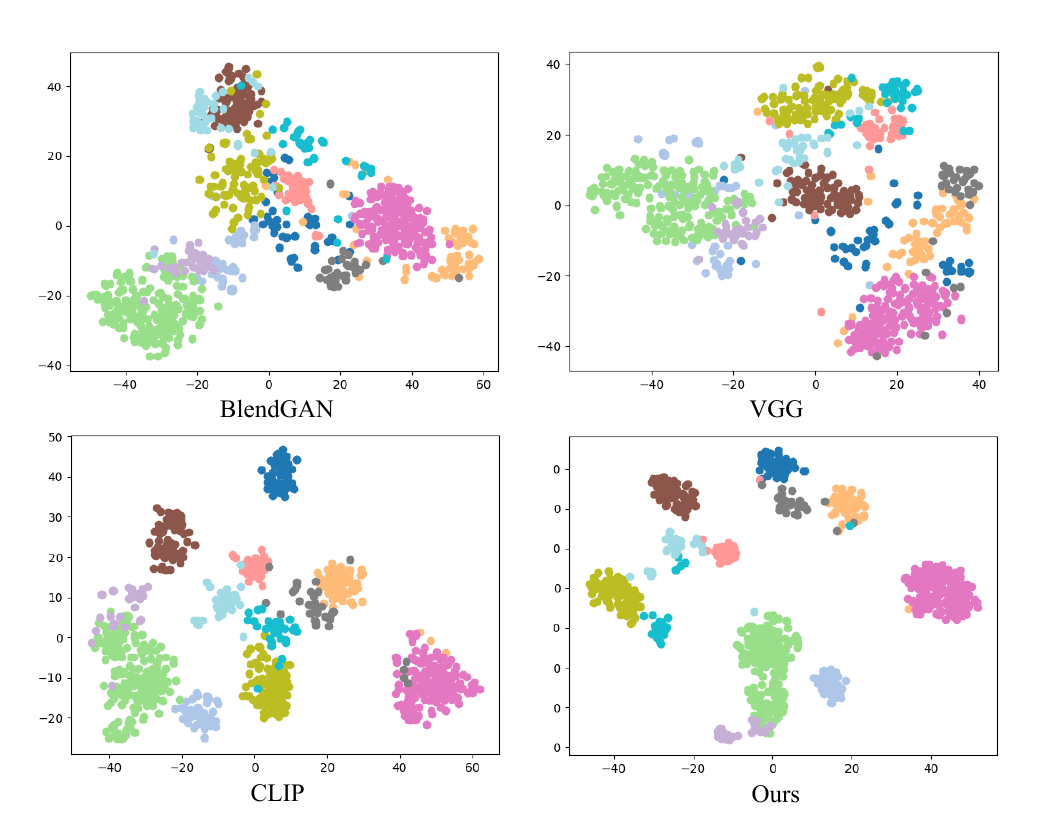}
	\caption{\textbf{Visualization of style embedding.} Different feature encoders are used to represent style images, and t-SNE is employed to visualize the results. We utilize different color markers to indicate the style categories of the images. Our Style Encoder exhibits better clustering characteristics for different style categories.} 
	\label{fig:cluster}
\end{figure}

\begin{table}[!b]
    \centering
    \caption{\textbf{Comparision with other feature extraction methods.} Silhouette Coefficient~\cite{rousseeuw1987silhouettes} is a measure of how similar an object is to its own cluster (cohesion) compared to other clusters (separation). Calinski-Harabasz index~\cite{calinski1974dendrite} is defined as the ratio of the sum of between-clusters dispersion and of within-cluster dispersion.}
    \label{tab:cluster}
    
    \setlength{\tabcolsep}{4mm}{
    \begin{tabularx}{\linewidth}{c|cccc}
        \toprule
        Methods & VGG~\cite{simonyan2014very} & BlendGAN~\cite{liu2021blendgan} &
        CLIP~\cite{radford2021learning} &
        Ours \\
        \midrule
        Silhouette$\uparrow$          & 0.039     & 0.129     & 0.168     & \textbf{0.308}   \\
        Calinski-Harabasz $\uparrow$     & 62.8     & 63.7   & 60.8     & \textbf{120.0}    \\
        \bottomrule
    \end{tabularx}}
\end{table}

We use different methods to extract the style embedding of each image in the validation set, and then visualize the distribution of these embeddings in the representation space, as shown in Fig.~\ref{fig:cluster}. Style embeddings belonging to different categories are represented by points of different colors and distributed in their own clusters at different locations in space. Our Style Encoder demonstrates the ability to effectively cluster images belonging to the same style category, resulting in compact intra-class distances and large inter-class differences. It indicates that our method has a better ability to present style from images and is robust enough to handle novel style variations.
Additionally, we perform quantitative evaluation by Silhouette Coefficient~\cite{rousseeuw1987silhouettes} and Calinski-Harabasz metrics~\cite{calinski1974dendrite} in Tab.~\ref{tab:cluster} used to evaluate 
the effect of clustering. Both visual results and cluster metrics demonstrate that our method effectively extracts features with better clustering results for images with the same styles compared with other extractors.

\begin{figure}[tb]
	\centering
	\includegraphics[width=0.9\textwidth]{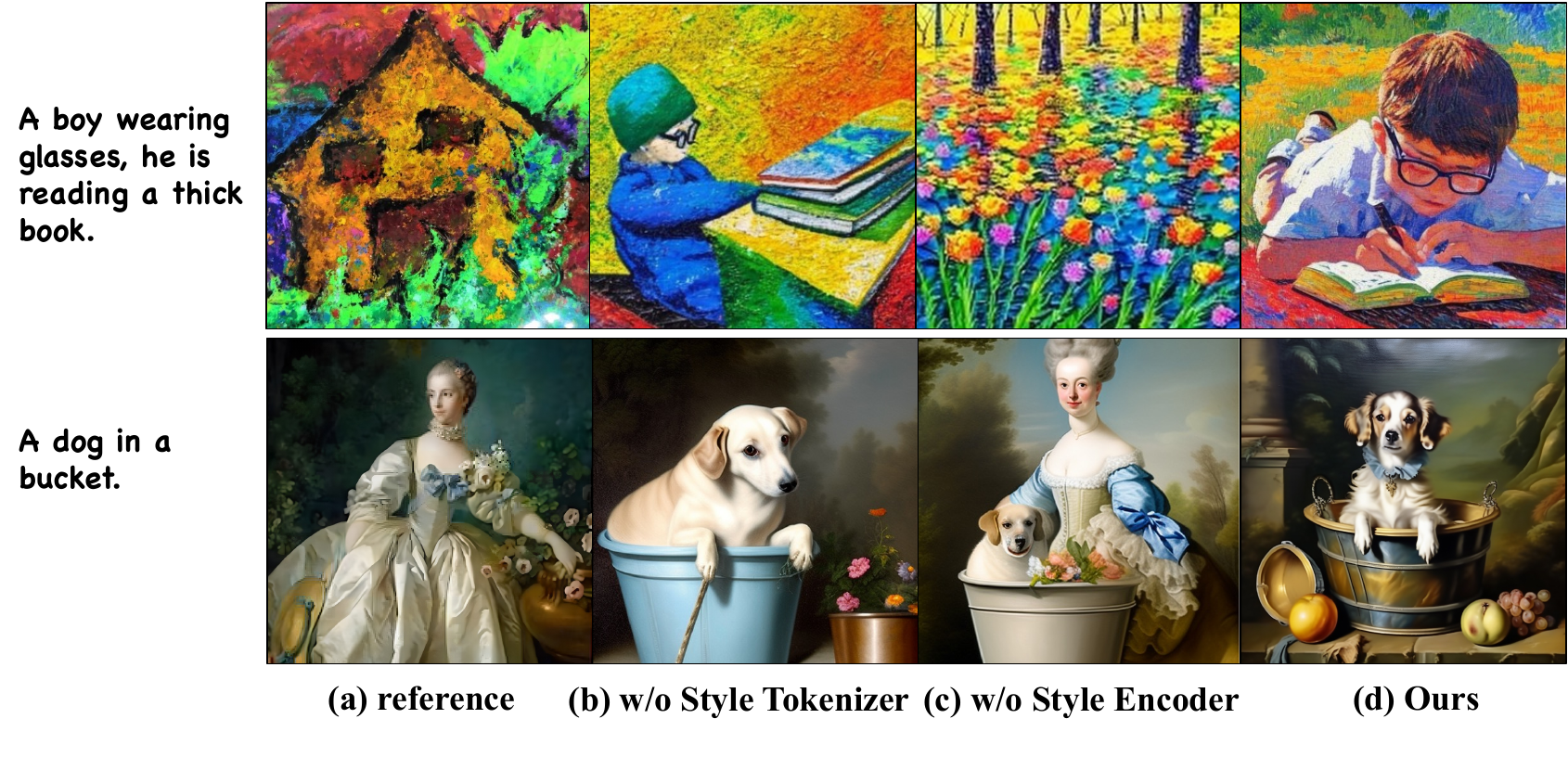}

	\caption{\textbf{The ablation study of our method.} \emph{w/o Style Tokenizer}: Style embedding is directly concatenated with the text embedding; \emph{w/o Style Encoder}: CLIP visual encoder is applied to extract image embedding.}
	\label{fig:ablation}
\end{figure}

\begin{table}[!b]
    \centering
    \caption{\textbf{Quantitative experiments of ablation study.} }
    \label{tab:quant_ablation}
    \setlength{\tabcolsep}{6mm}{
    \begin{tabularx}{\linewidth}{c|ccc}
    \hline
    Methods & Style-Sim     & Aesthetic     & \bf Text-Sim \\
    \hline
    w/o S.Tokenizer               & 0.468          & 6.43             & 0.215    \\
    w/o S.Encoder               & 0.475          & 6.62             & 0.127    \\
    \textbf{Ours}                     & \textbf{0.482}               & \textbf{6.68}    & \textbf{0.277}            \\
    \hline
    \end{tabularx}}
\end{table}

\subsection{Ablation Studies}
\label{sec:ablation}
In this section, we conduct ablation studies to assess the effectiveness of our Style Encoder and Style Tokenizer in Fig.~\ref{fig:ablation} and Table ~\ref{tab:quant_ablation}. Fig.~\ref{fig:ablation}(b) represents that style embedding is not first aligned to the word embedding space by the style tokenizer, but is directly concatenated with text embedding. 
Fig.~\ref{fig:ablation}(c) represents that the style encoder is not used for style representation, but the CLIP visual encoder is used directly to encode the image. 
Experimental results show that if any one of them is missing, the generated images either have a weakened ability to follow instructions or have poor style consistency.

\subsection{Other Application} 
\begin{figure*}[bt]
	\centering
	\includegraphics[width=0.9\textwidth]{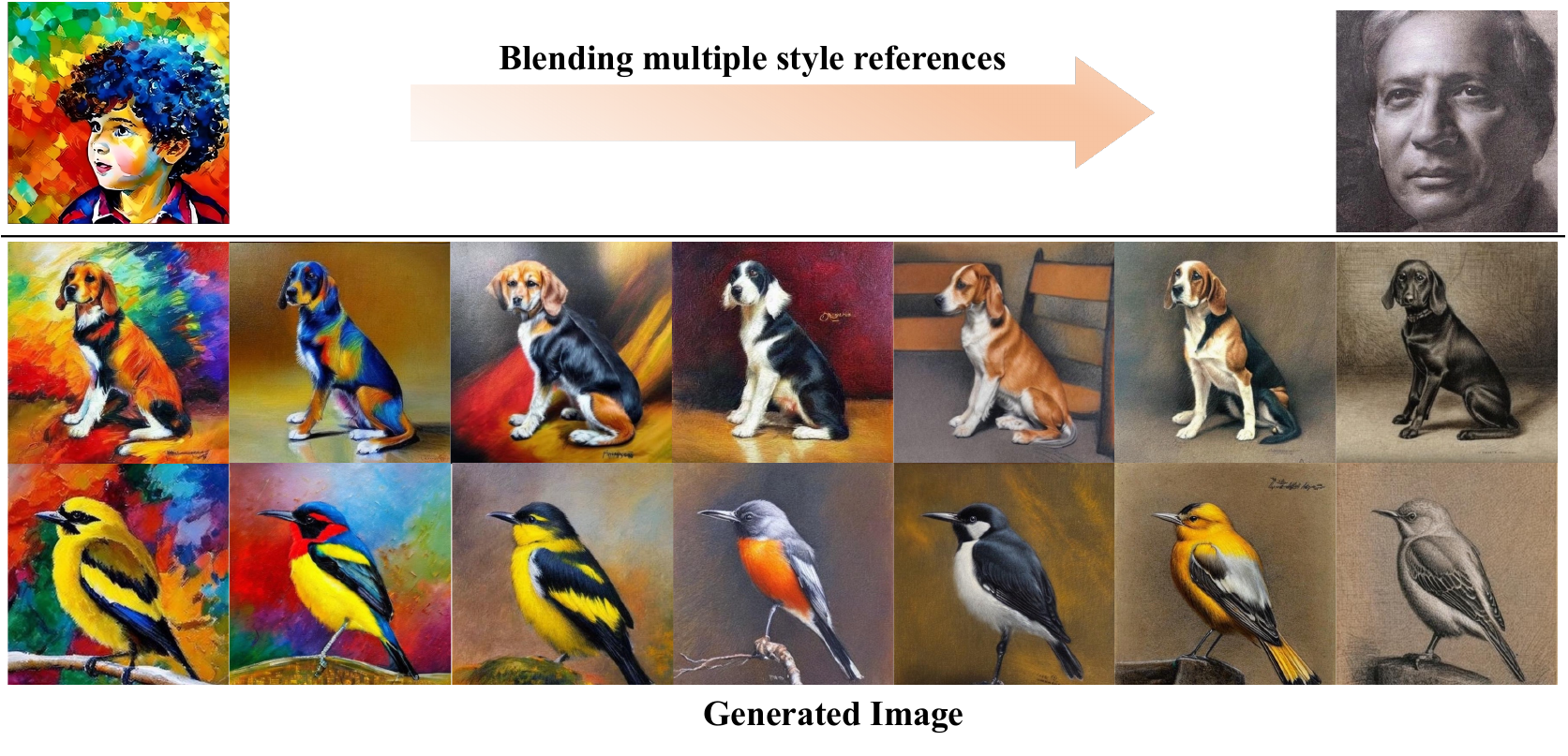}
	\caption{\textbf{Blend of different styles.} The first row is two reference images in different styles (Left: palette, Right: sketch). The second and third rows are style blend demonstrations whose text prompts are "a dog" and "a bird" respectively.} 
	\label{fig:blend}
\end{figure*}

Since our method can maintain the style in the reference image, if multiple images of different styles are used as references, the fusion between styles produces new styles. We use two styles for blending style in Fig.\ref{fig:blend}. By starting with the palette style as a control and gradually incorporating the sketch style, the images that are generated show a progressive transition from the palette to the sketch style. 

\section{Conclusion}
In this work, we propose a novel zero-shot method to precisely control the style of generated images from diffusion models. In order to decouple style and content conditions, we first construct a fine-labeled style dataset called Style30K and propose a Style Encoder that can extract style representation from reference images. Then we propose a Style Tokenizer to align style and text tokens in a uniform space. Finally, the aligned tokens are used as a condition in the denoising process of the diffusion model. Our method offers a flexible and effective solution for incorporating style control in image generation, opening up new possibilities for generating high-quality stylized content.


%
%
\bibliographystyle{splncs04}
\bibliography{main}

\begin{thebibliography}{10}
\providecommand{\url}[1]{\texttt{#1}}
\providecommand{\urlprefix}{URL }
\providecommand{\doi}[1]{https://doi.org/#1}

\bibitem{brooks2023instructpix2pix}
Brooks, T., Holynski, A., Efros, A.A.: Instructpix2pix: Learning to follow image editing instructions. In: Proceedings of the IEEE/CVF Conference on Computer Vision and Pattern Recognition. pp. 18392--18402 (2023)

\bibitem{brown2020language}
Brown, T., Mann, B., Ryder, N., Subbiah, M., Kaplan, J.D., Dhariwal, P., Neelakantan, A., Shyam, P., Sastry, G., Askell, A., et~al.: Language models are few-shot learners. Advances in neural information processing systems  \textbf{33},  1877--1901 (2020)

\bibitem{calinski1974dendrite}
Caliński, T., Harabasz, J.: A dendrite method for cluster analysis. Communications in Statistics  \textbf{3}(1),  1--27 (1974). \doi{10.1080/03610927408827101}

\bibitem{choi2018stargan}
Choi, Y., Choi, M., Kim, M., Ha, J.W., Kim, S., Choo, J.: Stargan: Unified generative adversarial networks for multi-domain image-to-image translation. In: Proceedings of the IEEE conference on computer vision and pattern recognition. pp. 8789--8797 (2018)

\bibitem{choi2020stargan}
Choi, Y., Uh, Y., Yoo, J., Ha, J.W.: Stargan v2: Diverse image synthesis for multiple domains. In: Proceedings of the IEEE/CVF conference on computer vision and pattern recognition. pp. 8188--8197 (2020)

\bibitem{deng2022stytr2}
Deng, Y., Tang, F., Dong, W., Ma, C., Pan, X., Wang, L., Xu, C.: Stytr2: Image style transfer with transformers. In: Proceedings of the IEEE/CVF conference on computer vision and pattern recognition. pp. 11326--11336 (2022)

\bibitem{gal2022image}
Gal, R., Alaluf, Y., Atzmon, Y., Patashnik, O., Bermano, A.H., Chechik, G., Cohen-Or, D.: An image is worth one word: Personalizing text-to-image generation using textual inversion. arXiv preprint arXiv:2208.01618  (2022)

\bibitem{gatys2015neural}
Gatys, L.A., Ecker, A.S., Bethge, M.: A neural algorithm of artistic style. arXiv preprint arXiv:1508.06576  (2015)

\bibitem{ho2022classifier}
Ho, J., Salimans, T.: Classifier-free diffusion guidance. arXiv preprint arXiv:2207.12598  (2022)

\bibitem{hu2021lora}
Hu, E.J., Shen, Y., Wallis, P., Allen-Zhu, Z., Li, Y., Wang, S., Wang, L., Chen, W.: Lora: Low-rank adaptation of large language models. arXiv preprint arXiv:2106.09685  (2021)

\bibitem{huang2023composer}
Huang, L., Chen, D., Liu, Y., Shen, Y., Zhao, D., Zhou, J.: Composer: Creative and controllable image synthesis with composable conditions. arXiv preprint arXiv:2302.09778  (2023)

\bibitem{huang2022diffstyler}
Huang, N., Zhang, Y., Tang, F., Ma, C., Huang, H., Zhang, Y., Dong, W., Xu, C.: Diffstyler: Controllable dual diffusion for text-driven image stylization. arXiv preprint arXiv:2211.10682  (2022)

\bibitem{huang2017arbitrary}
Huang, X., Belongie, S.: Arbitrary style transfer in real-time with adaptive instance normalization. In: Proceedings of the IEEE international conference on computer vision. pp. 1501--1510 (2017)

\bibitem{karras2020analyzing}
Karras, T., Laine, S., Aittala, M., Hellsten, J., Lehtinen, J., Aila, T.: Analyzing and improving the image quality of stylegan. In: Proceedings of the IEEE/CVF conference on computer vision and pattern recognition. pp. 8110--8119 (2020)

\bibitem{laionaesthetics}
{Laion.ai}: Laion-aesthetics. \url{https://laion.ai/blog/laion-aesthetics/}

\bibitem{li2023blip}
Li, D., Li, J., Hoi, S.C.: Blip-diffusion: Pre-trained subject representation for controllable text-to-image generation and editing. arXiv preprint arXiv:2305.14720  (2023)

\bibitem{liu2023llava}
Liu, H., Li, C., Wu, Q., Lee, Y.J.: Visual instruction tuning (2023)

\bibitem{liu2021blendgan}
Liu, M., Li, Q., Qin, Z., Zhang, G., Wan, P., Zheng, W.: Blendgan: Implicitly gan blending for arbitrary stylized face generation. Advances in Neural Information Processing Systems  \textbf{34},  29710--29722 (2021)

\bibitem{midjourney}
{MidJourney}: Midjourney. \url{https://www.midjourney.com/}

\bibitem{mou2023t2i}
Mou, C., Wang, X., Xie, L., Wu, Y., Zhang, J., Qi, Z., Shan, Y., Qie, X.: T2i-adapter: Learning adapters to dig out more controllable ability for text-to-image diffusion models. arXiv preprint arXiv:2302.08453  (2023)

\bibitem{dalle3}
OpenAI: Dall-e 3. \url{https://openai.com/dall-e-3}

\bibitem{podell2023sdxl}
Podell, D., English, Z., Lacey, K., Blattmann, A., Dockhorn, T., M{\"u}ller, J., Penna, J., Rombach, R.: Sdxl: Improving latent diffusion models for high-resolution image synthesis. arXiv preprint arXiv:2307.01952  (2023)

\bibitem{radford2021learning}
Radford, A., Kim, J.W., Hallacy, C., Ramesh, A., Goh, G., Agarwal, S., Sastry, G., Askell, A., Mishkin, P., Clark, J., et~al.: Learning transferable visual models from natural language supervision. In: International conference on machine learning. pp. 8748--8763. PMLR (2021)

\bibitem{ramesh2022hierarchical}
Ramesh, A., Dhariwal, P., Nichol, A., Chu, C., Chen, M.: Hierarchical text-conditional image generation with clip latents. arXiv preprint arXiv:2204.06125  \textbf{1}(2), ~3 (2022)

\bibitem{ramesh2021zero}
Ramesh, A., Pavlov, M., Goh, G., Gray, S., Voss, C., Radford, A., Chen, M., Sutskever, I.: Zero-shot text-to-image generation. In: International Conference on Machine Learning. pp. 8821--8831. PMLR (2021)

\bibitem{rombach2022high}
Rombach, R., Blattmann, A., Lorenz, D., Esser, P., Ommer, B.: High-resolution image synthesis with latent diffusion models. In: CVPR. pp. 10684--10695 (2022)

\bibitem{rousseeuw1987silhouettes}
Rousseeuw, P.J.: Silhouettes: a graphical aid to the interpretation and validation of cluster analysis. Computational and Applied Mathematics  \textbf{20},  53--65 (1987). \doi{10.1016/0377-0427(87)90125-7}

\bibitem{ruiz2023dreambooth}
Ruiz, N., Li, Y., Jampani, V., Pritch, Y., Rubinstein, M., Aberman, K.: Dreambooth: Fine tuning text-to-image diffusion models for subject-driven generation. In: Proceedings of the IEEE/CVF Conference on Computer Vision and Pattern Recognition. pp. 22500--22510 (2023)

\bibitem{ruta2021aladin}
Ruta, D., Motiian, S., Faieta, B., Lin, Z., Jin, H., Filipkowski, A., Gilbert, A., Collomosse, J.: Aladin: All layer adaptive instance normalization for fine-grained style similarity. In: Proceedings of the IEEE/CVF International Conference on Computer Vision. pp. 11926--11935 (2021)

\bibitem{saharia2022photorealistic}
Saharia, C., Chan, W., Saxena, S., Li, L., Whang, J., Denton, E.L., Ghasemipour, K., Gontijo~Lopes, R., Karagol~Ayan, B., Salimans, T., et~al.: Photorealistic text-to-image diffusion models with deep language understanding. Advances in Neural Information Processing Systems  \textbf{35},  36479--36494 (2022)

\bibitem{sauer2023stylegan}
Sauer, A., Karras, T., Laine, S., Geiger, A., Aila, T.: Stylegan-t: Unlocking the power of gans for fast large-scale text-to-image synthesis. arXiv preprint arXiv:2301.09515  (2023)

\bibitem{simonyan2014very}
Simonyan, K., Zisserman, A.: Very deep convolutional networks for large-scale image recognition. arXiv preprint arXiv:1409.1556  (2014)

\bibitem{voynov2023p+}
Voynov, A., Chu, Q., Cohen-Or, D., Aberman, K.: $ p+ $: Extended textual conditioning in text-to-image generation. arXiv preprint arXiv:2303.09522  (2023)

\bibitem{wang2024instantstyle}
Wang, H., Wang, Q., Bai, X., Qin, Z., Chen, A.: Instantstyle: Free lunch towards style-preserving in text-to-image generation. arXiv preprint arXiv:2404.02733  (2024)

\bibitem{wang2023cogvlm}
Wang, W., Lv, Q., Yu, W., Hong, W., Qi, J., Wang, Y., Ji, J., Yang, Z., Zhao, L., Song, X., Xu, J., Xu, B., Li, J., Dong, Y., Ding, M., Tang, J.: Cogvlm: Visual expert for pretrained language models (2023)

\bibitem{wang2023styleadapter}
Wang, Z., Wang, X., Xie, L., Qi, Z., Shan, Y., Wang, W., Luo, P.: Styleadapter: A single-pass lora-free model for stylized image generation. arXiv preprint arXiv:2309.01770  (2023)

\bibitem{wu2023not}
Wu, Y., Nakashima, Y., Garcia, N.: Not only generative art: Stable diffusion for content-style disentanglement in art analysis. In: Proceedings of the 2023 ACM International Conference on Multimedia Retrieval. pp. 199--208 (2023)

\bibitem{ye2023ip-adapter}
Ye, H., Zhang, J., Liu, S., Han, X., Yang, W.: Ip-adapter: Text compatible image prompt adapter for text-to-image diffusion models  (2023)

\bibitem{zhang2023adding}
Zhang, L., Rao, A., Agrawala, M.: Adding conditional control to text-to-image diffusion models. In: Proceedings of the IEEE/CVF International Conference on Computer Vision. pp. 3836--3847 (2023)

\bibitem{2023i2vgenxl}
Zhang, S., Wang, J., Zhang, Y., Zhao, K., Yuan, H., Qing, Z., Wang, X., Zhao, D., Zhou, J.: I2vgen-xl: High-quality image-to-video synthesis via cascaded diffusion models  (2023)

\bibitem{zhang2023prospect}
Zhang, Y., Dong, W., Tang, F., Huang, N., Huang, H., Ma, C., Lee, T.Y., Deussen, O., Xu, C.: Prospect: Expanded conditioning for the personalization of attribute-aware image generation. arXiv preprint arXiv:2305.16225  (2023)

\bibitem{zhang2023inversion}
Zhang, Y., Huang, N., Tang, F., Huang, H., Ma, C., Dong, W., Xu, C.: Inversion-based style transfer with diffusion models. In: Proceedings of the IEEE/CVF Conference on Computer Vision and Pattern Recognition. pp. 10146--10156 (2023)

\bibitem{zhang2022domain}
Zhang, Y., Tang, F., Dong, W., Huang, H., Ma, C., Lee, T.Y., Xu, C.: Domain enhanced arbitrary image style transfer via contrastive learning. In: ACM SIGGRAPH 2022 Conference Proceedings. pp.~1--8 (2022)

\bibitem{zhu2017unpaired}
Zhu, J.Y., Park, T., Isola, P., Efros, A.A.: Unpaired image-to-image translation using cycle-consistent adversarial networks. In: Proceedings of the IEEE international conference on computer vision. pp. 2223--2232 (2017)

\end{thebibliography}

\end{document}



\noindent
\large \textbf{Appendix}

\setcounter{page}{1}
\setcounter{page}{1}
\setcounter{section}{1}
\renewcommand\thesection{\Alph{section}}

\subsection{Dataset for Evaluation}
\label{sec:sup_dataset}
\begin{figure}[!h]
	\centering
	\includegraphics[width=1\textwidth]{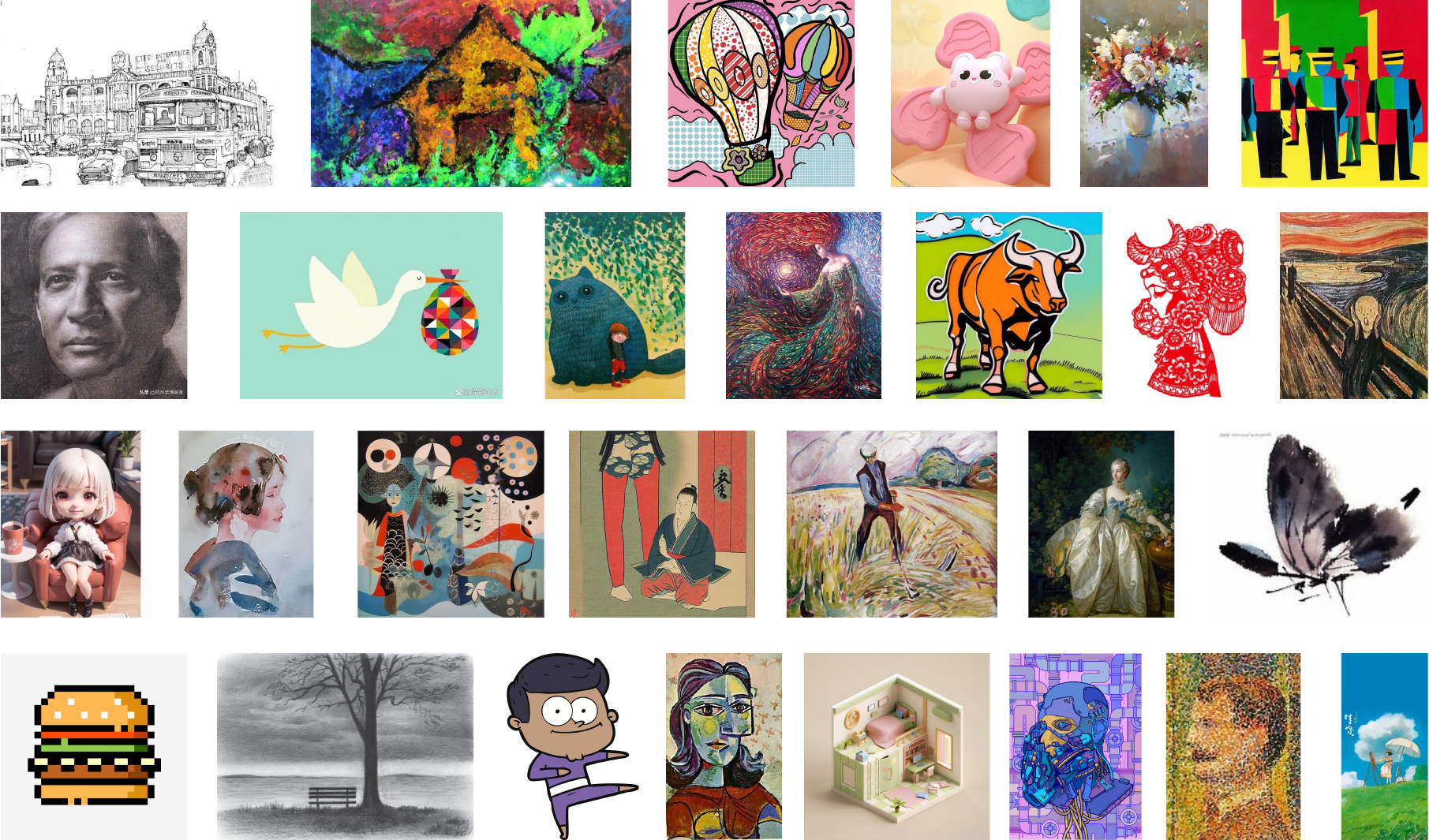}
	\caption{\textbf{The 28 style reference images with diverse styles for the generation models.}}
	\label{fig:sup_styleref}
\end{figure}

In the experiment, we use a combination of 52 prompts and 28 style images, resulting in 1456 unique combinations, to guide the generation of images. These images are then used for quantitative evaluation.
The list of prompts is shown in Tab.~\ref{tab:sup_prompt}, using the same setting as in StyleAdapter~\cite{wang2023styleadapter}. The 28 style reference images are selected from the validation references to cover a wide range of styles, displayed in Fig.~\ref{fig:sup_styleref}.

\begin{longtable}{l}
\caption{\textbf{The list of prompts for experiments.}}
\label{tab:sup_prompt}\\

\toprule
\textbf{Prompt} \\
\midrule
\endfirsthead

A robot.  \\
A girl wearing a red dress, she is dancing.  \\
A boy wearing glasses, he is reading a thick book.  \\ 
A little cute boy.  \\
A woman wearing a green sportswear, she is running.  \\ 
A woman wearing a purple hat and a yellow scarf.  \\ 
A man wearing a black leather jacket and a red tie.  \\ 
A little boy with glasses and a watch.  \\
A smiling little girl.  \\
A little boy playing football.  \\
A curly-haired boy.  \\
A little girl holding flowers.  \\
A lovely kitten walking in a garden.  \\
A puppy sitting on a sofa.  \\
A fluffy white rabbit with pink ears and nose.  \\
A brown puppy with black spots and a red collar.  \\ 
A black and white panda.  \\
A dog in a bucket.  \\
A cat wearing a hat.  \\
A cute little fish in aquarium.  \\
A bird in a word.  \\
A kitten sleeping on a pillow.  \\
A parrot singing a song.  \\
A monkey playing with a banana.  \\
A turtle wearing sunglasses.  \\
A hamster eating a carrot.  \\
A white rose.  \\
A sunflower smiling at the sun.  \\
A cactus wearing a hat.  \\
A daisy with a ladybug on it.  \\
A pine tree with a snowman hugging it.  \\
A mushroom in winter.  \\
A beautiful lotus.  \\
A lotus with a frog meditating on it.  \\
A cherry blossom.  \\
A palm tree.  \\
A river with rapids and rocks.  \\
A creek with clear water and colorful pebbles.  \\
A lake with calm water and reflections.  \\
A waterfall with mist and rainbows.  \\
A stone with a face carved on it, standing on a pedestal in a museum.  \\ 
A stone with a hole in it.  \\
A stone with a pattern of stripes on it.  \\
A stone with a crack in it, holding a plant growing out of it.  \\
A snowy mountain peak.  \\
A mountain goat on a cliff.  \\
A red baseball cap.  \\
A football on the grass.  \\
A motorcycle.  \\
A modern house with a pool.  \\
A house made of cardboard boxes.  \\
A house covered with ice and snow.  \\

\bottomrule
\end{longtable}

\subsection{Ablation study of hyper-parameter.}

\begin{figure}[!b]
  \centering
  \includegraphics[width=1.0\linewidth]{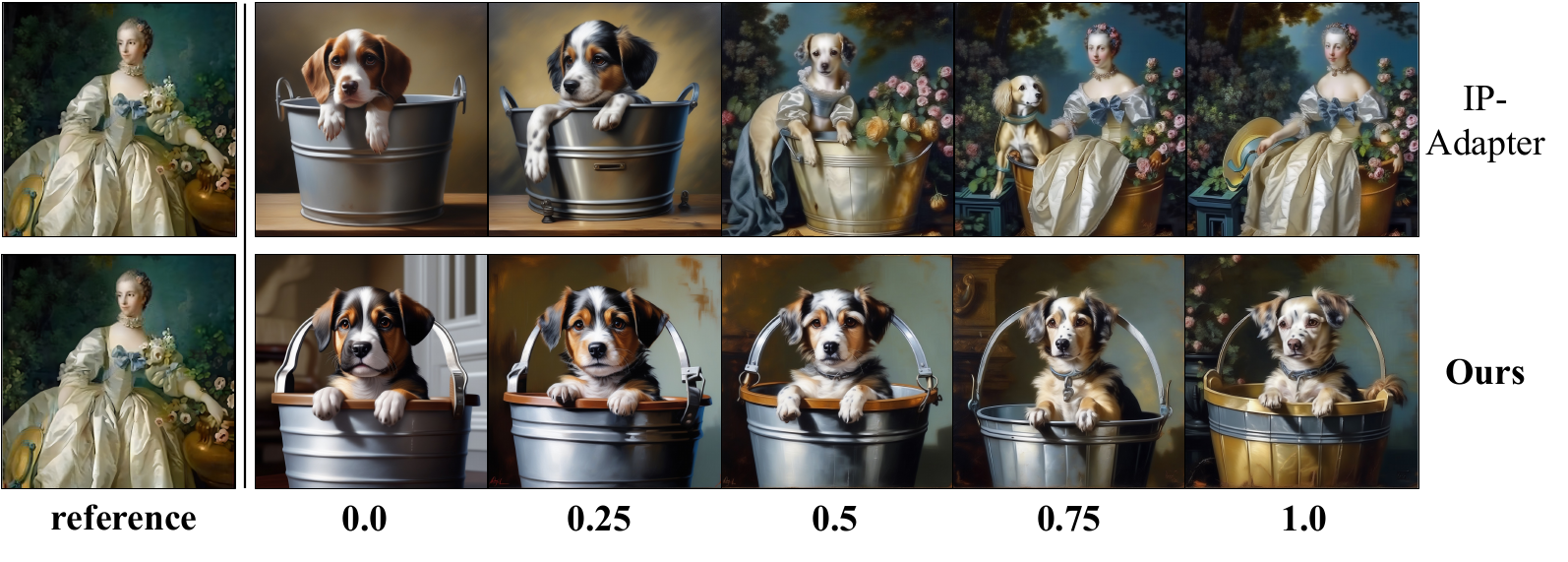}
   \caption{\textbf{Ablation study of hyper-parameter.}}
   \label{fig:hyper_param}
\end{figure}

To verify the stability of the trade-off between instruction following and style control of this method, we evaluate our method and IP-Adapter under different control weights, respectively. As shown in Fig.\ref{fig:hyper_param}, by applying different style control weights, the semantics of the generated images remain unchanged. And a higher weight has a greater impact on the style of the image.

\subsection{Visualization of Style Similarity Score}

\begin{figure}[!h]
	\centering
	\includegraphics[width=1\textwidth]{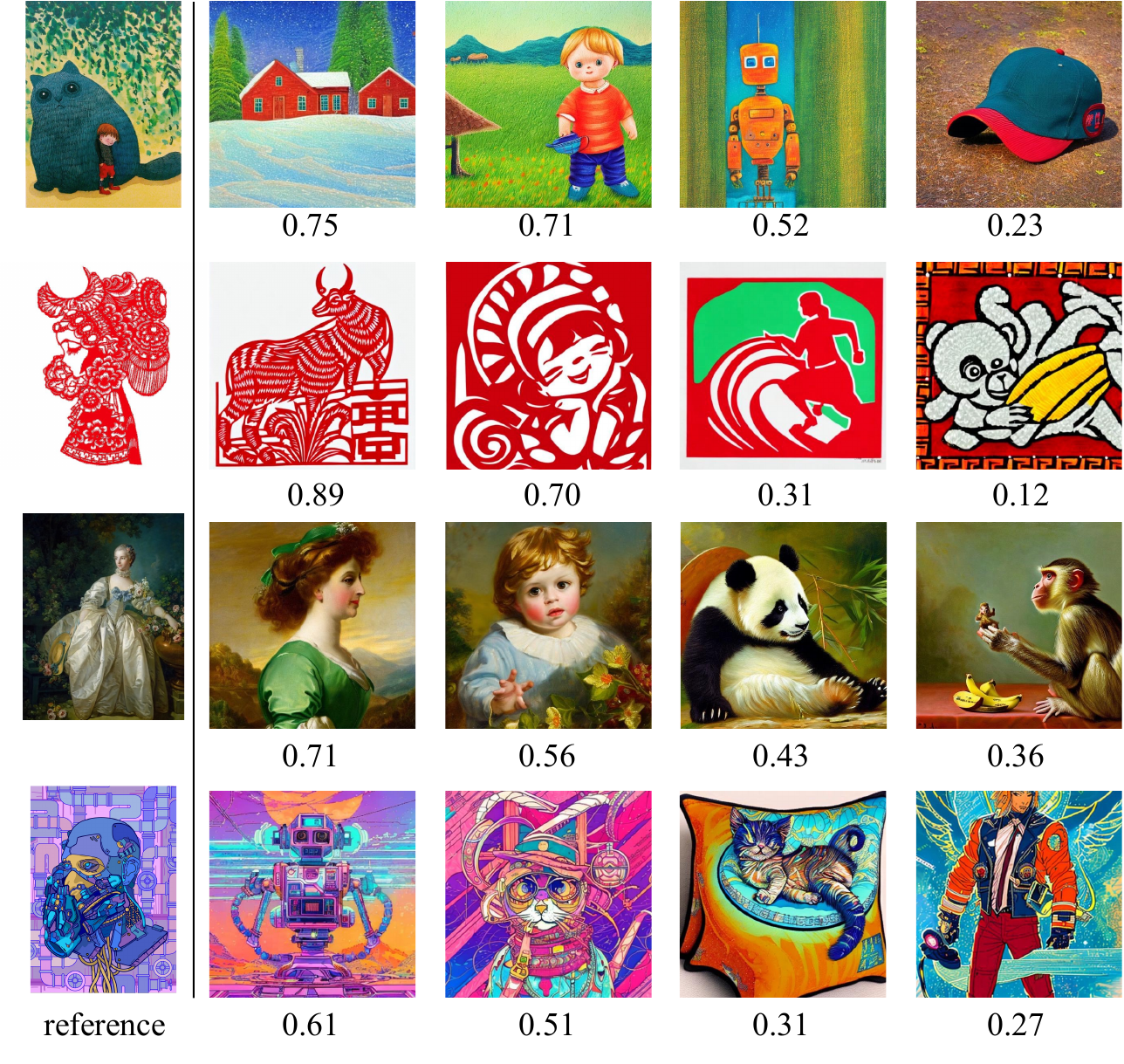}
	
	\caption{\textbf{Comparison of style similar scores between different pairs of style images and reference images.}}
	\label{fig:sup_stylescore}
\end{figure}

In the experiment, we employ a style encoder to measure the style similarity between two images. In Fig.~\ref{fig:sup_stylescore}, we present different generated images and annotate their style similarity scores compared with their respective style reference images. It can be observed that the style encoder accurately assesses the similarity of styles between two images. Moreover, images with a style similarity score greater than 0.4 exhibit a favorable level of style similarity.

\subsection{More Generated Images With More Styles}
\begin{figure*}[tb]
	\centering
	\includegraphics[width=\textwidth]{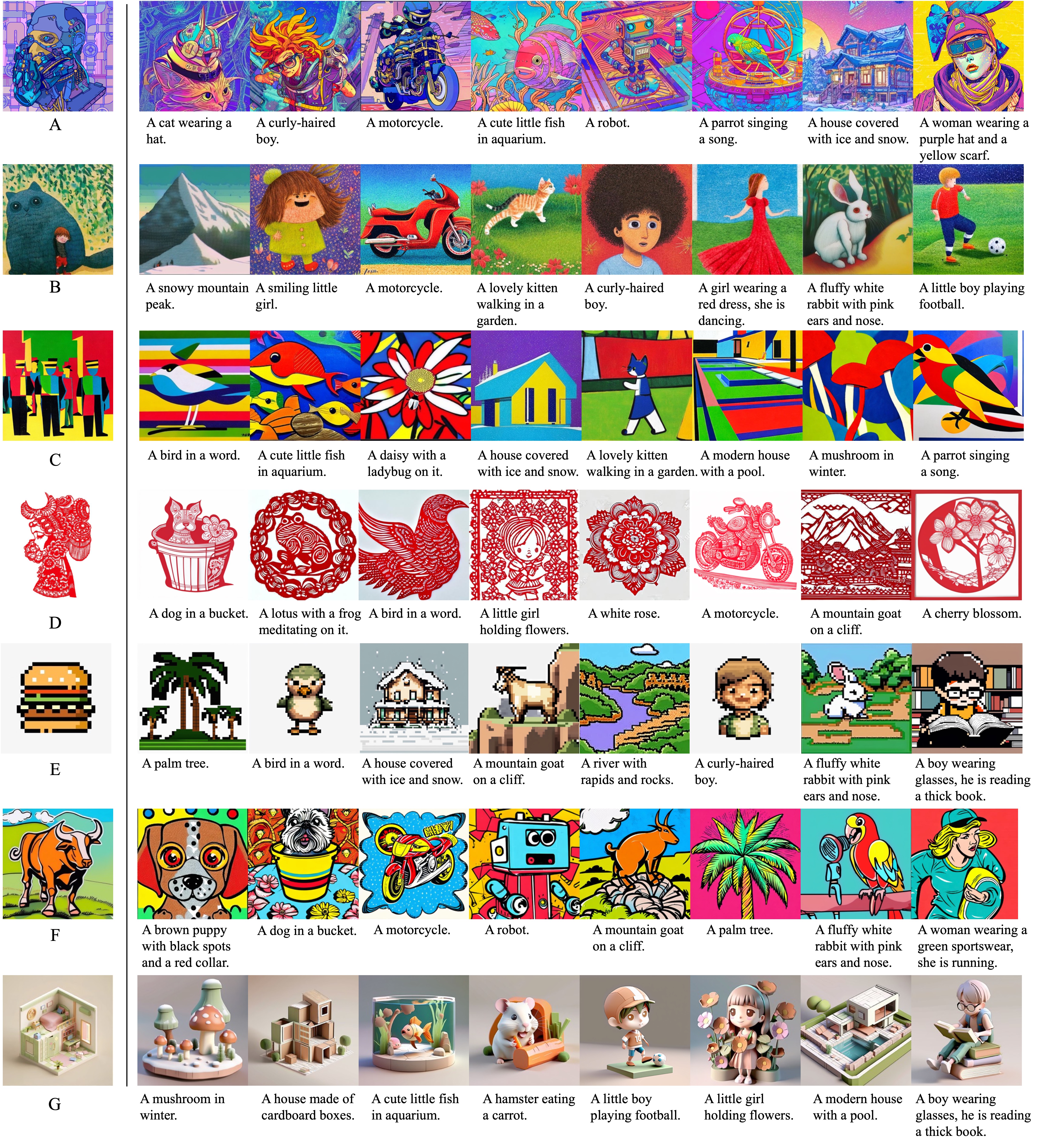}
	
	\caption{\textbf{Additional visual results generated by our algorithm.} The first image on the left of each row is the corresponding style reference image, followed by the prompt and its corresponding generated results.}
	\label{fig:sup_gen}
\end{figure*}

In Fig.~\ref{fig:sup_gen}, we present additional generated results. The first image in each row represents the style reference image, followed by the corresponding generated results. The prompts for each image are listed below the images. It can be observed that our method can generate images in various styles. 


\clearpage  

%
%
\bibliographystyle{splncs04}
\bibliography{main}